\documentclass{article}

\usepackage[preprint]{neurips_2025}

\usepackage[utf8]{inputenc} 
\usepackage[T1]{fontenc}    
\usepackage{hyperref}       
\usepackage{url}            
\usepackage{booktabs}       
\usepackage{amsfonts}       
\usepackage{nicefrac}       
\usepackage{microtype}      
\usepackage{xcolor}         
\usepackage{graphicx}       
\usepackage{amsmath}        
\usepackage{amssymb}        

\title{Graph Neural Networks vs Convolutional Neural Networks for Graph Domination Number Prediction}

\author{%
  Randy Davila \\ 
  Department of Computational Applied Mathematics \& Operations Research \\
  Rice University \\
  Houston, USA \\
  \texttt{rrd6@rice.edu} \\
  \And
  Beyzanur Ispir \\ 
  Department of Computational Applied Mathematics \& Operations Research \\
  Rice University \\
  Houston, USA \\
  \texttt{beyza.ispir@rice.edu} \\
}

\begin{document}

\maketitle

\begin{abstract}
We investigate machine learning approaches to approximating the \emph{domination number} of graphs, the minimum size of a dominating set. Exact computation of this parameter is NP-hard, restricting classical methods to small instances. We compare two neural paradigms: Convolutional Neural Networks (CNNs), which operate on adjacency matrix representations, and Graph Neural Networks (GNNs), which learn directly from graph structure through message passing. Across 2,000 random graphs with up to 64 vertices, GNNs achieve markedly higher accuracy ($R^2=0.987$, MAE $=0.372$) than CNNs ($R^2=0.955$, MAE $=0.500$). Both models offer substantial speedups over exact solvers, with GNNs delivering more than $200\times$ acceleration while retaining near-perfect fidelity. Our results position GNNs as a practical surrogate for combinatorial graph invariants, with implications for scalable graph optimization and mathematical discovery.
\end{abstract}

\section{Introduction}\label{sec:introduction}
Graphs provide a unifying mathematical framework for modeling complex systems, from communication and transportation networks to molecular structures and biological interactions. Among the many invariants studied in graph theory, the \emph{domination number} $\gamma(G)$ plays a central role in problems of resource allocation, coverage, and network security. A dominating set is a subset of vertices $D$ such that every vertex outside $D$ has a neighbor in $D$; the domination number is the minimum size of such a set. Foundational work in the 1970s established the basis of domination theory \cite{Cockayne1977,Allan1978}, with comprehensive treatments consolidating its importance in structural graph theory \cite{Haynes1998}. However, computing $\gamma(G)$ is NP-complete \cite{Garey1979}, and even approximation guarantees are constrained by hardness-of-approximation results \cite{Feige1998}. Classical heuristics \cite{Parekh1991} and refined exact algorithms \cite{Fomin2009} remain restricted to small instances, underscoring the need for scalable alternatives.

One emerging approach is to replace algorithmic computation with prediction. Rather than solving each instance of domination from scratch, a machine learning model can be trained to map graph structure directly to $\gamma(G)$. The idea of ``learning to optimize'' has gained traction across combinatorial domains \cite{Khalil2017}, with recent work demonstrating that neural models can approximate graph invariants such as stability numbers \cite{Davila2024}. Domain-specific tools such as \texttt{GraphCalc} \cite{Davila2025} now enable systematic experimentation with these surrogates, opening the door to large-scale studies that would be infeasible with classical solvers alone.

At the architectural level, the natural candidates are Graph Neural Networks (GNNs). Originating from early formulations of graph-based recurrent networks \cite{Scarselli2009}, the field has rapidly advanced through gated updates \cite{Li2016}, graph convolutions \cite{Kipf2017}, inductive methods \cite{Hamilton2017}, and general message passing \cite{Gilmer2017}. Of particular relevance is the Graph Isomorphism Network (GIN) \cite{Xu2019}, which matches the expressive power of the Weisfeiler--Lehman test \cite{Morris2019} and has become a standard choice for learning structural graph properties.

As a contrasting paradigm, Convolutional Neural Networks (CNNs) have been adapted to graphs by treating adjacency matrices as images. This approach leverages convolutional filters to capture local connectivity \cite{Tixier2018} and, in certain settings, has even rivaled or outperformed graph-based models \cite{Boronina2023}. Yet CNNs discard the permutation invariance intrinsic to graphs, raising questions about their suitability for learning combinatorial invariants that depend on global graph structure.

Against this backdrop, we present a comparative study of CNNs and GNNs for predicting the domination number. By situating CNNs as a vision-inspired baseline and GNNs as a graph-native model, we quantify how architectural inductive biases translate into predictive accuracy, runtime efficiency, and robustness across graph sizes. To our knowledge, this is the first systematic study of domination number prediction within this comparative framework, and our findings highlight the promise of GNNs as scalable surrogates for hard graph invariants.

\section{Related Work}\label{sec:related-work}
The study of domination in graphs has a long history. Foundational papers established key structural results \cite{Cockayne1977,Allan1978}, and comprehensive surveys \cite{Haynes1998} underline its centrality in graph theory and applications. From a computational standpoint, exact algorithms remain limited due to NP-completeness \cite{Garey1979} and hardness of approximation \cite{Feige1998}, though specialized heuristics \cite{Parekh1991} and exponential-time algorithms \cite{Fomin2009} have been developed. These barriers motivate surrogate approaches capable of scaling beyond traditional methods.

Graph Neural Networks have emerged as a powerful paradigm for learning on graph-structured data. Early formulations of graph-based neural computation \cite{Scarselli2009,Li2016} evolved into modern message-passing architectures \cite{Kipf2017,Hamilton2017,Gilmer2017}, with the Graph Isomorphism Network (GIN) providing near-optimal expressive power \cite{Xu2019,Morris2019}. GNNs have been successfully applied across domains, from chemistry to combinatorial optimization \cite{Khalil2017}, and have shown promise in approximating graph invariants \cite{Davila2024}. Tools like \texttt{GraphCalc} \cite{Davila2025} now provide systematic infrastructure for these investigations.

Convolutional Neural Networks, while originally developed for vision, have been adapted to graphs by operating on adjacency matrices. This line of work exploits convolutional locality \cite{Tixier2018} and has even produced competitive baselines for graph classification \cite{Boronina2023}. However, CNNs inherently discard permutation invariance, raising doubts about their ability to capture invariants such as domination number that depend on graph isomorphism classes.

Our work bridges these strands by directly comparing GNNs and CNNs on the task of predicting domination numbers. While prior studies have demonstrated the potential of neural surrogates for combinatorial problems, a focused evaluation of domination---a classical yet computationally challenging parameter---has not been carried out. By filling this gap, we provide both empirical benchmarks and conceptual insights into how architectural inductive biases shape performance on hard graph invariants.

\section{Data representation and network design}\label{sec:design}
Our study frames the domination number as a regression target: given a graph $G$, the goal is to predict $\gamma(G)$ as closely as possible to the exact value computed by \texttt{GraphCalc}. The dataset consists of 2{,}000 Erd\H{o}s--R\'enyi graphs $G(n,p)$ with $n$ ranging from 5 to 64 and edge probability $p$ drawn uniformly from $[0,1]$. To address real-world applicability concerns, we also evaluate on 2{,}000 Barabasi--Albert graphs $G(n,m)$ with $n \in [5,64]$ and $m = 2$, which model scale-free networks found in social networks, internet topology, and biological systems. Each graph is labeled with its exact domination number, and the collection is split into 80\% training and 20\% testing, stratified by graph size.

To establish a baseline using standard vision architectures, we represent graphs as images. The adjacency matrix of each graph is augmented with degree information along the diagonal, rescaled to the range $[0,255]$, and padded or resized to $64 \times 64$ to accommodate varying graph orders. A convolutional network is then applied:
\begin{align}
\text{Conv2D}(32, 3 \times 3) &\;\rightarrow\; \text{MaxPool}(2 \times 2) \\
\text{Conv2D}(64, 3 \times 3) &\;\rightarrow\; \text{MaxPool}(2 \times 2) \\
\text{Conv2D}(64, 3 \times 3) &\;\rightarrow\; \text{Flatten} \\
\text{Dense}(64) &\;\rightarrow\; \text{Dense}(1).
\end{align}
This architecture leverages locality priors of convolution but discards the permutation invariance intrinsic to graphs.

For the graph-native approach, we employ a three-layer Graph Isomorphism Network (GIN). Each layer updates node embeddings through sum aggregation of neighbor features, followed by a small multilayer perceptron (MLP) with ReLU activation and batch normalization:
\begin{align}
h_v^{(l+1)} &= \text{MLP}^{(l)}\!\left((1 + \epsilon^{(l)}) \cdot h_v^{(l)} + \sum_{u \in \mathcal{N}(v)} h_u^{(l)}\right), \\
\text{where} \quad \text{MLP}^{(l)} &= \text{Linear} \circ \text{ReLU} \circ \text{Linear}.
\end{align}
After three rounds of message passing, a graph-level representation is formed by concatenating mean and additive pooling over node embeddings:
\begin{equation}
h_G = \text{Concat}\big[\text{MeanPool}(H), \; \text{AddPool}(H)\big],
\end{equation}
which is then passed through a linear layer to predict $\gamma(G)$. Note that $\epsilon^{(l)}$ is a learnable parameter.

Both models are trained with the Adam optimizer using a learning rate of $10^{-3}$ and mean squared error loss. The CNN converges within roughly 25 epochs, while the GNN is trained for up to 200 epochs with gradient clipping and early stopping on a validation split. Hyperparameters such as hidden dimension and pooling strategy were tuned by grid search. Performance is evaluated using mean absolute error (MAE), root mean squared error (RMSE), and coefficient of determination ($R^2$), complemented by runtime experiments on 64-vertex graphs to assess speedup relative to exact computation. All implementations and experiments can be found in a companion Colab notebook\footnote{
\url{https://colab.research.google.com/drive/1biduFhO8cDs_PSe0abGYlSH8hUBeSacd?usp=sharing}}.

\section{Results}\label{sec:results}

We evaluate both accuracy and efficiency on the held-out test set of 400 Erd\H{o}s--R\'enyi graphs. Table~\ref{tab:results} shows a clear advantage for the GNN: it achieves an $R^2$ of 0.987 compared to 0.955 for the CNN, corresponding to a 26\% reduction in MAE and a 47\% reduction in RMSE. These findings confirm that message-passing architectures are better suited to capturing structural dependencies than convolution applied to adjacency matrices.

\begin{table}[h]
  \caption{Prediction accuracy on ER test set (400 graphs).}
  \label{tab:results}
  \centering
  \begin{tabular}{lccc}
    \toprule
    Method & MAE & RMSE & $R^2$ \\
    \midrule
    GNN & 0.372 & 0.463 & 0.987 \\
    CNN & 0.500 & 0.874 & 0.955 \\
    \bottomrule
  \end{tabular}
\end{table}

Efficiency is equally important, since exact solvers scale poorly with graph size. Table~\ref{tab:runtime} compares prediction times on graphs with 64 vertices. While the CNN achieves a modest $7.4\times$ speedup, the GNN is over two orders of magnitude faster, requiring only 1.1 milliseconds per instance. This combination of accuracy and runtime efficiency makes GNNs attractive for real-time applications where exact methods are impractical.

\begin{table}[h]
  \caption{Average runtime on 64-vertex graphs (20 trials).}
  \label{tab:runtime}
  \centering
  \begin{tabular}{lcc}
    \toprule
    Method & Avg. Time (ms) & Speedup \\
    \midrule
    GraphCalc (Exact) & 234.9 & 1.0$\times$ \\
    CNN & 31.6 & 7.4$\times$ \\
    GNN & 1.1 & 208.9$\times$ \\
    \bottomrule
  \end{tabular}
\end{table}

We also examine the role of architectural choices within the GNN. As shown in Table~\ref{tab:ablation}, mean pooling alone performs poorly (MAE $=1.571$), whereas concatenating mean and additive pooling reduces error by more than 76\%. This highlights the importance of combining distributional (mean) and cumulative (sum) information when aggregating node embeddings.

\begin{table}[h]
  \caption{Pooling strategies for the GNN.}
  \label{tab:ablation}
  \centering
  \begin{tabular}{lc}
    \toprule
    Pooling Strategy & MAE \\
    \midrule
    Mean + Add Pooling & 0.372 \\
    Mean Pooling Only & 1.571 \\
    \bottomrule
  \end{tabular}
\end{table}

Performance also scales well with the number of vertices in the graph. Table~\ref{tab:detailed_results} reports the results by vertex ranges. GNNs maintain $R^2 > 0.97$ even on the largest graphs, while CNNs degrade more noticeably as size increases. This suggests that permutation-invariant message passing provides a more robust foundation for learning combinatorial properties across scales.

\begin{table}[h]
  \caption{Prediction accuracy by graph size ranges (MAE for GNN and CNN, and $R^2$ for GNN).}
  \label{tab:detailed_results}
  \centering
  \begin{tabular}{lccc}
    \toprule
    Graph Size & GNN MAE & CNN MAE & GNN $R^2$ \\
    \midrule
    5--20 vertices & 0.245 & 0.512 & 0.991 \\
    21--40 vertices & 0.387 & 0.698 & 0.981 \\
    41--64 vertices & 0.518 & 0.871 & 0.975 \\
    \bottomrule
  \end{tabular}
\end{table}

Finally, to assess real-world applicability, we trained both models on Barab\'asi--Albert (BA) graphs, which model scale-free networks in social, technological, and biological systems. The results were consistent with those of the ER graphs: the GNN achieved $R^2 = 0.981$ (MAE = 0.395, RMSE = 0.496), while the CNN lagged $R^2 = 0.908$ (MAE = 0.797, RMSE = 1.095). Both models performed slightly worse than on the ER graphs, reflecting the added complexity of scale-free structures, but the GNN maintained a stable 7--8\% advantage. However, cross-domain testing revealed limited generalization: ER-trained models performed poorly on BA graphs ($R^2 = -0.237$ for GNN, $0.473$ for CNN), underscoring the importance of domain-specific training. All experiments are available in a companion Colab notebook\footnote{
\url{https://colab.research.google.com/drive/1biduFhO8cDs_PSe0abGYlSH8hUBeSacd?usp=sharing}}.





\section{Conclusion and Future Work}\label{sec:conclusion}

We compared convolutional and graph-based neural architectures for predicting the domination number, a classical yet hard graph invariant. Graph neural networks consistently achieved higher accuracy and more than two orders of magnitude speedup over exact solvers, while CNNs on adjacency images provided a weaker but informative baseline. These results confirm that inductive biases aligned with graph structure are decisive for learning combinatorial properties.

Looking ahead, several extensions are natural. The same framework can be applied to other invariants—such as independence, chromatic, or zero forcing numbers—where exact methods remain infeasible, and where neural surrogates could serve as heuristics or pruning rules in optimization pipelines. An intriguing direction is to invert the predictive model: using trained weights to guide the generation of graphs with specified domination numbers. Such generative capabilities would not only broaden practical applications but also open new avenues for exploring the structure of hard graph invariants.

\bibliographystyle{plainnat}  
\bibliography{refs}
\end{document}